\documentclass[10pt,twocolumn,letterpaper]{article}

\usepackage{cvpr}              
\usepackage{booktabs}
\usepackage{multirow}
\usepackage{array}
\usepackage{listings}
\usepackage{xcolor}
\usepackage{graphicx}
\usepackage[most]{tcolorbox}
\usepackage{pgf-pie}
\usepackage{tikz}
\usetikzlibrary{arrows.meta}
\usetikzlibrary{shapes.geometric, positioning}
\usepackage{subcaption}
\usepackage{float}
\usepackage{adjustbox}
\usepackage{graphicx}

\newtcolorbox{conversationbox}[2][]{%
    colback=white,
    colframe=black!70,
    boxrule=1pt,
    title=\textbf{#2},
    left=1mm, 
    right=1mm,
    top=1mm, 
    bottom=1mm,
    before skip=5pt,
    after skip=0pt,
    breakable=false,
    sharp corners=south, 
    #1
}

\newtcolorbox{subconversationbox}[1][]{%
    colback=white,
    colframe=black!70,
    boxrule=1pt,
    left=1mm, 
    right=1mm,
    top=1mm, 
    bottom=1mm,
    before skip=0pt, 
    after skip=5pt,
    breakable=false,
    sharp corners=north,
    #1
}

\definecolor{cvprblue}{rgb}{0.21,0.49,0.74}
\usepackage[pagebackref,breaklinks,colorlinks,allcolors=cvprblue]{hyperref}

\def\paperID{10} 
\def\confName{CVPR}
\def\confYear{2025}

\title{Understanding and Mitigating Toxicity in Image-Text Pretraining Datasets: A Case Study on LLaVA}

\author{
Karthik Reddy Kanjula\thanks{Equal contribution.} $^1$,
Surya Guthikonda\footnotemark[1] $^{3,1}$,
Nahid Alam\thanks{Work does not belong to position referred in $^2$.} $^{2,1}$, 
Shayekh Bin Islam $^{4,1}$ \\
\\
$^1$Cohere for AI Community, 
$^2$Cisco Meraki, 
$^3$Indiana University Bloomington, \\
$^4$Bangladesh University of Engineering and Technology \\
{\tt\small karthikreddykanjula99@gmail.com}
}

\begin{document}
\maketitle
\begin{abstract}
Pretraining datasets are foundational to the development of multimodal models, yet they often have inherent biases and toxic content from the web-scale corpora they are sourced from. In this paper, we investigate the prevalence of toxicity in LLaVA image-text pretraining dataset, examining how harmful content manifests in different modalities. We present a comprehensive analysis of common toxicity categories and propose targeted mitigation strategies, resulting in the creation of a refined toxicity-mitigated dataset. This dataset removes 7,531 of toxic image-text pairs in the LLaVA pre-training dataset. We offer guidelines for implementing robust toxicity detection pipelines. Our findings underscore the need to actively identify and filter toxic content - such as hate speech, explicit imagery, and targeted harassment - to build more responsible and equitable multimodal systems. The toxicity-mitigated dataset is open source and is available for further research.
\end{abstract}    
\section{Introduction}
\label{sec:intro}

Vision Language Models (VLMs) have emerged as a key paradigm in artificial intelligence, enabling machines to understand and reason about the visual world through natural language. Recent advances in large language models (LLMs) and general-purpose image encoders such as CLIP \citep{radford2021learning} and SigLIP \citep{zhai2023sigmoid} have significantly boosted VLM capabilities. Architectures such as Flamingo \citep{alayrac2022flamingo}, LLaVA \citep{liu2023llava, liu2023improvedllava}, KOSMOS \citep{kosmos-g, peng2023kosmos}, Florence-2 \citep{xiao2024florence} and Molmo \citep{deitke2024molmo} demonstrate strong performance across tasks including image captioning, Visual Question Answering (VQA), and complex reasoning. Qwen2-VL \citep{wang2024qwen2} introduced Multimodal Rotary Position Embedding (M-RoPE) \citep{su2021roformer} and dynamic resolution techniques, while PaLI’s joint modality scaling and cross-lingual learning \citep{chen2022pali, chen2023pali} have furthered vision-language understanding. LLaVA \citep{liu2023llava} architecture demonstrates mechanisms for processing visual tokens, enabling it to develop interpretable representations of visual content that align with textual semantics \citep{neo2024interpretingvisualinformationprocessing}. Despite these advances, there is limited research in understanding the toxic content of the image-text pretraining dataset.

VLMs are trained on billions of image-text pairs from diverse sources, such as LAION-5B \citep{schuhmann2022laion5bopenlargescaledataset} with 5.85 billion CLIP-filtered samples. While large-scale, web-scraped datasets enable robust training, they often contain toxic content that harms model performance and introduces ethical risks \citep{singla2024pixelsproselargedataset}. Models like LLaVA  \citep{liu2023llava}, which rely on these datasets, can inherit toxic or biased material, posing challenges for safe, responsible AI development.

Current VLM datasets can contain toxic and culturally insensitive content \cite{yue2024pangea}, including hate speech, explicit violence, sexual material, harmful stereotypes, and discriminatory representations. This content may appear in both visual and textual descriptions \citep{chen2024commcoherentinterleavedimagetext}. Recent "jailbreaking" techniques \citep{jin2024jailbreakzoosurveylandscapeshorizons} underscore the urgency of mitigating these risks. These issues highlight the need for careful examination to ensure a responsible and ethical deployment of vision-language models.

In this paper, we describe a multimodal approach to detect and remove harmful content from the LLaVA \citep{liu2023llava}  pretrain dataset. By combining multiple filters with large pretrained VLMs, we remove harmful visual and textual content while preserving the dataset’s richness and diversity. This approach produces a toxicity-mitigated dataset that minimizes societal risks and promotes responsible AI development.

Our key contributions include:
\begin{enumerate}
	\item Applied Toxic-BERT to flag 892 image captions as toxic with over 80\% confidence.
            \item Combined findings from LlavaGuard \cite{helff2024llavaguard} and Command R+ \citep{command_r} (7,111 images) with Toxic-BERT’s \citep{Detoxify} results (892 images) to identify a total of 7,531 unique toxic images.
            \item In total, we removed the 7,531 toxic images from the pretraining dataset, resulting in a toxicity-mitigated version for LLaVA pretraining

\end{enumerate}

\section{Related Work}
\label{sec:relatedwork}

In recent years, the LLaVA architecture and its associated datasets \citep{liu2023llava, liu2023improvedllava} have been widely adopted VLM tasks. However, to the best of our knowledge, the presence of toxic content in the LLaVA \citep{liu2023llava} pretraining dataset has not yet been systematically analyzed. In the context of VLMs, toxicity refers to harmful or offensive content that may emerge from biases or aggressive language present in the training data. Since these models are often trained on large-scale datasets that can contain historical prejudices or explicit material, such toxicity may surface in unpredictable and concerning ways.

\subsection{Toxicity Filtering in Dataset}

SPA-VL \citep{zhang2025spavlcomprehensivesafetypreference} proposes a Safety Preference Alignment dataset for Vision Language Models named SPA-VL. In terms of breadth, SPA-VL covers 6 harmfulness domains, 13 categories, and 53 subcategories, and contains 100,788 samples of the quadruple (question, image, chosen response, rejected response). ELITE \citep{lee2025eliteenhancedlanguageimagetoxicity} shows that current VLMs can incorporate self-filtering mechanisms to remove toxic content from training datasets. Chen et al. \citep{chen2024visionlanguagemodelstrongfilter} propose a self-filtering method. The authors demonstrate that VLMs themselves can serve as a filter for instruction-finetuning. Singla et al. \citep{singla2024pixelsproselargedataset} introduce PixelProse. To ensure data integrity, the authors rigorously analyze the dataset for problematic content, including child sexual abuse material (CSAM), personally identifiable information (PII), and toxicity using Google APIs.

\subsection{Safety Alignment In Model Architecture}

Zhao et al. \citep{zhao2025zeroshotdefensetoxicimages} propose harmful content mitigation techniques without relying on pre-filtering or fine-tuning techniques that incur higher cost. Instead, they introduce SafeCLIP, a lightweight method that leverages LVLMs inherent multimodal alignment for zero-shot toxic image detection. By projecting CLIPs discarded CLS token into its text space and matching it with toxic descriptors, SafeCLIP detects harmful content without any architectural changes-adding minimal latency and enabling dynamic safety corrections during inference.  Zou et al. \citep{zou2025understandingrectifyingsafetyperception} demonstrate that the inclusion of visual modality shifts activations toward a safer direction, which is a key factor contributing to the degradation of safety alignment. The authors propose ShiftDC, a method for disentangling and calibrating VLM activations to restore safety alignment.

\subsection{Evaluating Toxicity}
MM-SafetyBench \citep{liu2024mmsafetybenchbenchmarksafetyevaluation} shows that VLMs safety can be breached if the input image contains relevant harmful images related to the text prompt. To address this, MM-SafetyBench \citep{liu2024mmsafetybenchbenchmarksafetyevaluation} proposes a framework designed to conduct safety-critical evaluations of VLMs against such image-based manipulations. VHELM \citep{lee2024vhelmholisticevaluationvision} proposes a comprehensive framework for evaluating VLMs on biases, fairness, safety and toxicity. ELITE \citep{lee2025eliteenhancedlanguageimagetoxicity} evaluator explicitly incorporates a toxicity score to accurately assess harmfulness in multimodal contexts, where VLMs often provide specific, convincing, but unharmful descriptions of images. ELITE filters out ambiguous and low-quality image-text pairs from existing benchmarks using the ELITE evaluator and generates diverse combinations of safe and unsafe image-text pairs.


\section{Methodology}

\subsection{Dataset Toxicity Analysis}

LLaVA Pretrain dataset has 558,000 image-caption pairs in English language. Upon manually identifying the types of images present in the LLaVA Pretrain dataset by cherry picking, we discovered some toxic images. This led us to design a systematic approach for detecting and mitigating harmful content. To methodically identify and filter the toxic content that exists in both modalities, we used two different models. LlavaGuard 7B \cite{helff2024llavaguard} for images and Toxic-BERT \cite{Detoxify} for text. Our choice was guided by their specialized capabilities and performance in detecting harmful content within their respective modalities. LlavaGuard 7B was selected for its ability to identify and classify toxic visuals based on predefined safety guidelines. This ensures a comprehensive coverage of harmful visual elements. We chose Toxic-BERT for text because of its strong ability to identify offensive and harmful language through contextual analysis, which outperforms conventional keyword-based filtering methods. These models provided a reliable foundation for systematically identifying, understanding, and mitigating toxicity within the LLaVA Pretrain dataset.

\begin{figure}[t]
  \centering
  \fbox{\includegraphics[width=1\linewidth]{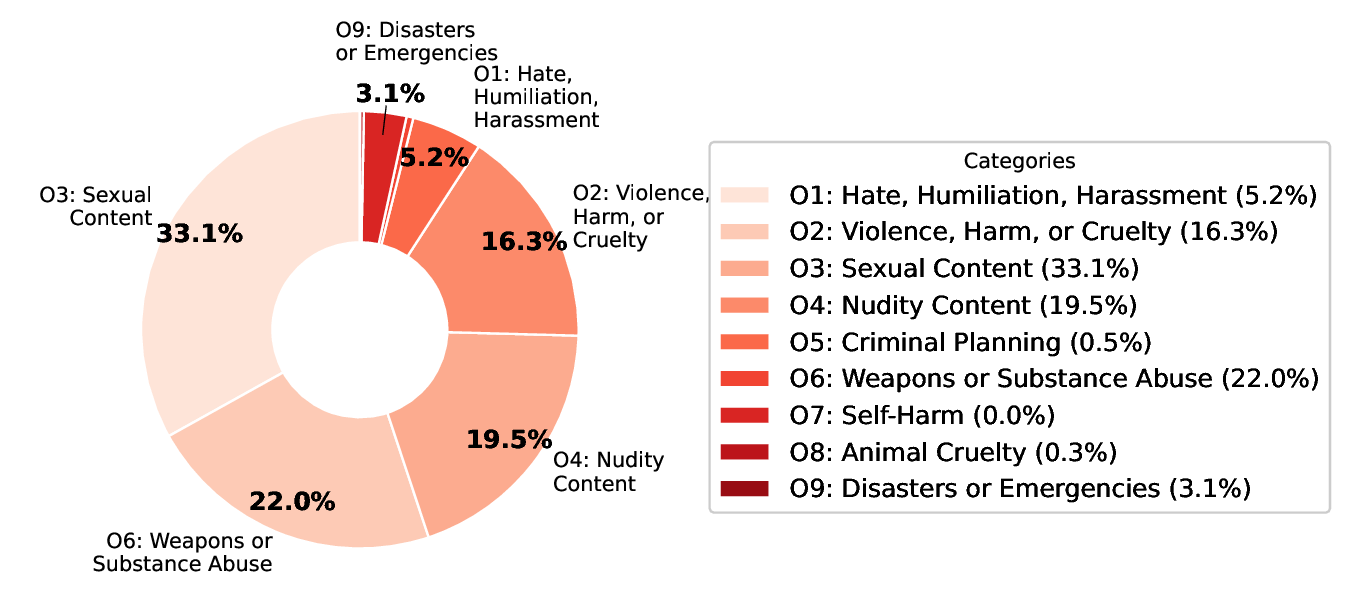}}
  \caption{Image Toxicity Analysis on LLaVA Pre-train Dataset using LlavaGuard}
  \label{fig:toxanalysis}
\end{figure}

The LLaVA Guard safety taxonomy presents a guideline for assessing visual content safety across nine distinct policy categories: Hate/Harassment (O1), Violence (O2), Sexual Content (O3), Nudity (O4), Criminal Planning (O5), Weapons/Substance Abuse (O6), Self-Harm (O7), Animal Cruelty (O8), and Disasters/Emergencies (O9). Each individual category interprets prohibited content (such as hateful speech, explicit nudity, or content promoting violence), also allowing for educational, informational, or artistic use cases within appropriate contexts. This assessment process requires evaluators to rate content as "Safe" or "Unsafe", assign the relevant policy category, and provide a detailed rationale on why a specific policy description is chosen. This process creates a structured approach to identify and filter potentially harmful content in LLaVA pretrain image datasets, as visualized in the toxicity analysis chart (Figure \ref{fig:toxanalysis}). In addition, a prompt optimization pipeline was introduced to ensure precise filtering - capturing potentially harmful content while reducing false positives.

\begin{figure}[t]
  \centering
  \fbox{\includegraphics[width=1\linewidth]{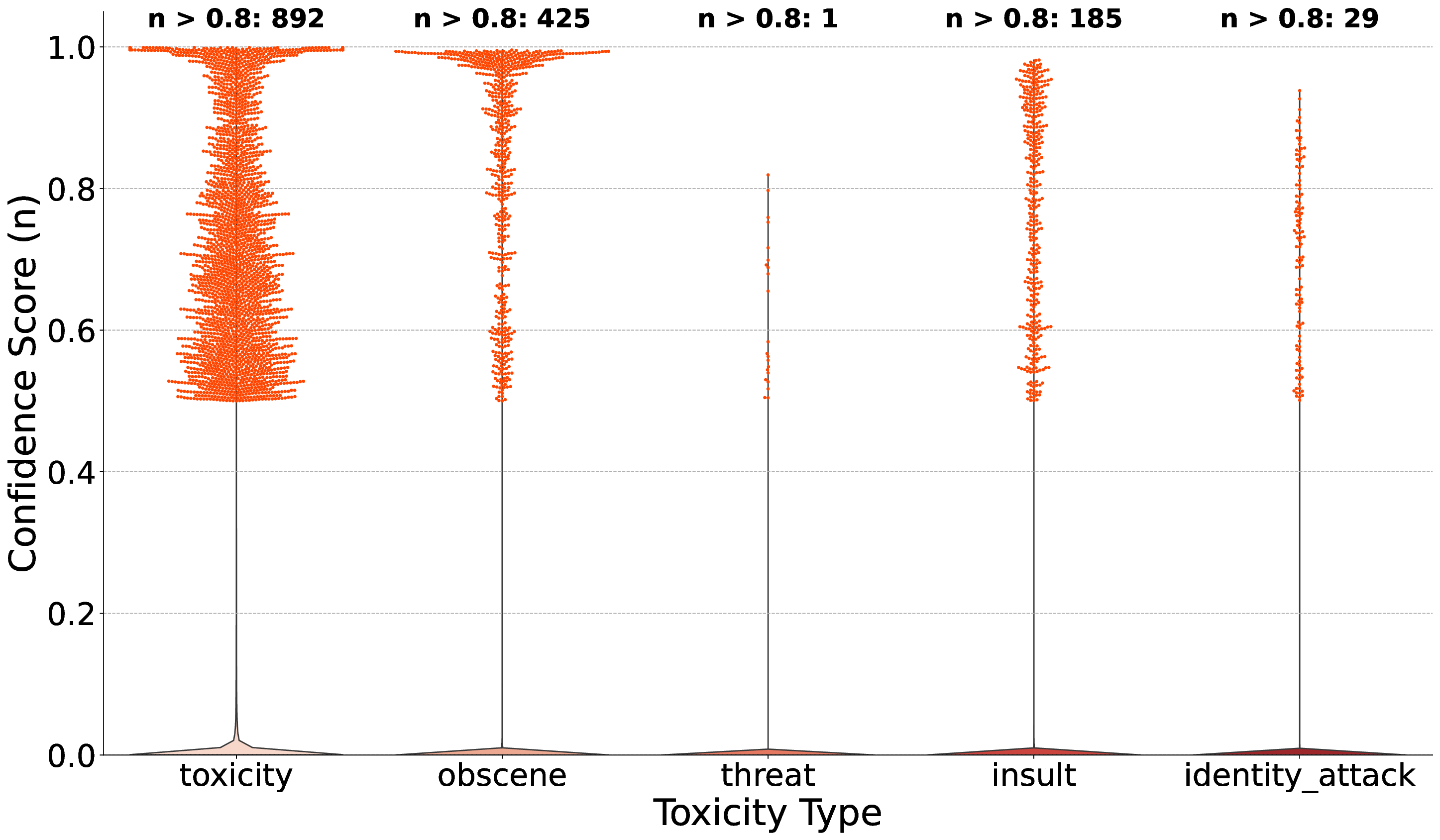}}
  \caption{Image Caption Toxicity Analysis on LLaVA Pre-train Dataset using Toxic-BERT}
  \label{fig:toxbert}
\end{figure}

\subsection{Dataset Toxicity Filtering}
\begin{figure}[t]
  \centering
  \fbox{\includegraphics[width=\linewidth]{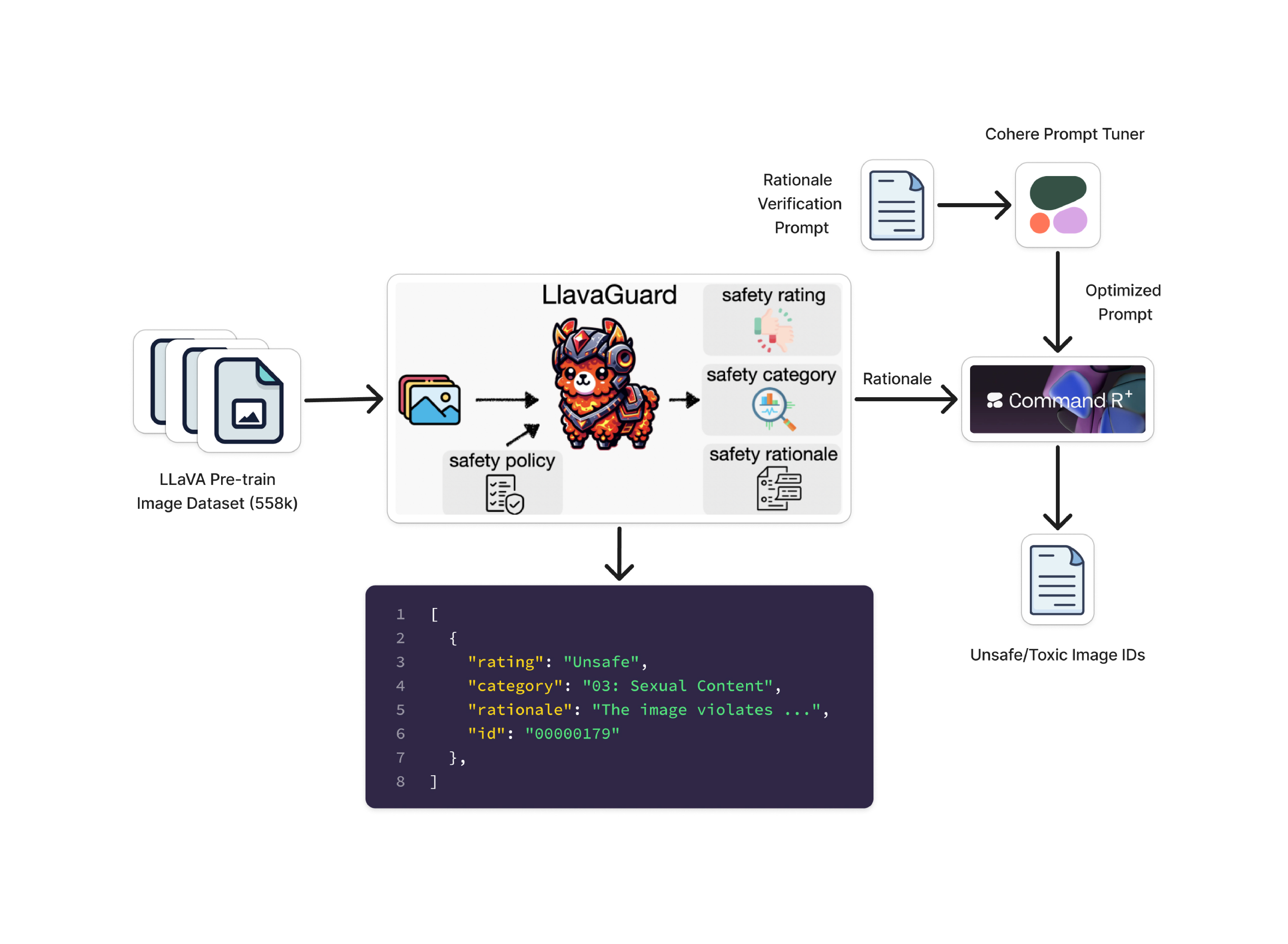}}
  \caption{Dataset Toxicity Filtering Method}
  \label{fig:toxfilter}
\end{figure}

The overall process for creating the toxicity-mitigated dataset is shown in Figure \ref{fig:toxfilter}. LlavaGuard output provides a rating, category, and rationale explaining why an image violates the guidelines. We then refine this by identifying genuinely toxic images. This is done by developing an optimized prompt using the Cohere prompt tuner\footnote{\url{https://docs.cohere.com/v2/docs/prompt-tuner}}. We use this prompt as a preamble (System Prompt) and pair it with the LlavaGuard output of each image flagged as unsafe. Command R+ \cite{command_r} then analyzes these results to identify the truly unsafe image IDs. In our analysis, LlavaGuard identified 7,600 images as toxic and the final Command R+ output identified  7,111 images as unsafe.

To detect harmful language in image captions, we utilized Toxic-BERT, a model fine-tuned on the Jigsaw Toxic Comment Classification Challenge dataset. This model is adept at identifying various forms of toxicity, including threats, obscenity, and identity-based hate. Its contextual understanding allows for nuanced detection beyond simple keyword matching, effectively capturing offensive language that might be context-dependent. Applying Toxic-BERT to our dataset, we identified 892 image captions as toxic with a confidence level that exceeds 80\%, as depicted in Figure \ref{fig:toxbert}. LlavaGuard and Command R+ identified 7,111 images and Toxic-BERT identified 892 images; there are in total 7,531 unique toxic images. We then removed those 7,531 images from the pretraining dataset to create the toxicity-mitigated pre-train dataset.

\section{Results}
\label{sec:results}

In this work, we create a toxicity-mitigated version of the LLaVA image-text pre-training dataset. We applied Toxic-BERT to flag 892 image captions as toxic with over 80\% confidence. We then combined the findings of LlavaGuard and Command R+ (7,111 images) with Toxic-BERT’s results (892 images) to identify a total of 7,531 unique toxic images. In total, we removed 7,531 toxic image-text pairs from the pretraining dataset, resulting in a toxicity-mitigated version for LLaVA pretraining. This data set is open source and is available to everyone for further research. 

\section{Future Work}
\label{sec:futurework}

Our research in multimodal safety is advanced through dataset toxicity mitigation. It would be interesting to verify the accuracy of the toxicity-mitigated dataset by applying a user evaluation process or through other toxicity-mitigation pipelines. In future we want to implement safety at instruction tuning and alignment techniques using SPA-VL’s safety preference alignment \citep{zhang2025spavlcomprehensivesafetypreference} during instruction tuning—with its domain-specific harmfulness taxonomies and automated labeling—to better define safety boundaries; adopting SafeCLIP \citep{zhao2025zeroshotdefensetoxicimages} inspired dynamic safety projection for efficient, real-time toxic detection; and employing capability-preserving model merging from Howard et al. \citep{ratzlaff2024trainingfreemitigationlanguagereasoning} to sustain reasoning and visual performance.
\section{Conclusion}
\label{sec:conclusion}

In this work, we introduced a fully open-source, toxicity-mitigated version of the LLaVA image-text pretraining dataset, offering a valuable resource for the broader research community. Future evaluations will leverage established benchmarks such as MM-SafetyBench \citep{liu2024mmsafetybenchbenchmarksafetyevaluation}, VHELM \citep{lee2024vhelmholisticevaluationvision}, and ELITE \citep{lee2025eliteenhancedlanguageimagetoxicity} to assess performance, robustness, and safety. We hope this effort encourages researchers to prioritize the qualitative aspects of data—particularly toxic or harmful content—in their endeavors, ultimately fostering more responsible and ethical AI development.
{
    \small
    \bibliographystyle{ieeenat_fullname}
    \bibliography{main}
}


\end{document}